\documentclass[11pt, letterpaper]{article}
\usepackage{geometry} 

\usepackage{subfigure}
\usepackage{epstopdf}
\usepackage{epsfig}
\usepackage{array}
\usepackage{lipsum}
\usepackage[]{algorithm2e}
\usepackage{tabularx}
\usepackage{hyperref}

\usepackage{enumitem}

\usepackage{booktabs} 

\usepackage{graphicx}
\usepackage{booktabs}
\usepackage{multirow}
\usepackage{multicol}
\usepackage{array}
\usepackage{tabularx}
\usepackage{bm}
\usepackage{soul}
\usepackage{tikz}
\usepackage{amssymb}
\usepackage{amsmath}
\usepackage{xfrac}
\usepackage[bottom]{footmisc}
\usepackage{fixltx2e}

\def\invcircledast#1{%
  \mathbin{\vphantom{\circledast}\text{%
    \ooalign{\smash{\blackcircle}\cr
             \hidewidth\smash{\textcolor{white}{\bf \footnotesize $#1$}}\hidewidth\cr
            }%
  }}%
}

\newcommand{\blackcircle}{\raisebox{-.6ex}{\scalebox{2.30}{$\bullet$}}}

\newcommand{\Design}{$\mathsf{SpikeHD}$\xspace}
\newcommand{\Designn}{SpikeHD\xspace}

\usepackage{etoolbox}

\usepackage{bm}

\usepackage[numbers,sort&compress,square]{natbib}

\date{}
\begin{document}

\title{\textbf{Spiking Hyperdimensional Network: Neuromorphic Models Integrated with Memory-Inspired Framework}}

\author{\bf  Zhuowen Zou$^1$, Haleh Alimohamadi$^2$, Farhad Imani$^{3}$, Yeseong Kim$^4$ \\[3pt] \textbf{Mohsen Imani}$^{5*}$}
\date{\small 
$^1$University of California San Diego, La Jolla, CA 92093\\
$^2$University of California Los Angeles, Los Angeles, CA 90095\\
$^3$ University of Connecticut, Storrs, CT 06269 \\
$^4$ Daegu Gyeongbuk Institute of Science and Technology, South Korea \\
$^5$University of California Irvine, Irvine, CA 92697 \\
\medskip 
\medskip 
$^*$To whom correspondence should be addressed. E-mail: m.imani@uci.edu \\

\date{} 
\vspace{-3mm}}

\medskip 

\maketitle
\pagestyle{plain}

\begin{abstract}
Recently, brain-inspired computing models have shown great potential to outperform today's deep learning solutions in terms of robustness and energy efficiency. Particularly, Spiking Neural Networks (SNNs) and HyperDimensional Computing (HDC) have shown promising results in enabling efficient and robust cognitive learning. Despite the success, these two brain-inspired models have different strengths. While SNN mimics the physical properties of the human brain, HDC models the brain on a more abstract and functional level. Their design philosophies demonstrate complementary patterns that motivate their combination. With the help of the classical psychological model on memory, we propose \Designn, the first framework that fundamentally combines Spiking neural network and hyperdimensional computing. \Designn generates a scalable and strong cognitive learning system that better mimics brain functionality.  \Designn exploits spiking neural networks to extract low-level features by preserving the spatial and temporal correlation of raw event-based spike data. Then, it utilizes HDC to operate over SNN output by mapping the signal into high-dimensional space, learning the abstract information, and classifying the data. 
Our extensive evaluation on a set of benchmark classification problems shows that \Designn provides the following benefit compared to SNN architecture: (1) significantly enhance learning capability by exploiting two-stage information processing, (2) enables substantial robustness to noise and failure, and (3) reduces the network size and required parameters to learn complex information. 

\end{abstract}


\section{Introduction}
Many applications run machine learning algorithms to assimilate the data collected in the swarm of devices on the Internet of Things (IoT). 
Sending all the data to the cloud for processing is not scalable, cannot guarantee a real-time response. 
However, the high computational complexity and memory requirement of existing DNNs hinder usability to a wide variety of real-life embedded applications where the device resources and power budget is limited~\cite{denil2013predicting, zaslavsky2013sensing, sun2016internet, xiang2019pipelined}. Therefore, we need alternative learning methods to train on the less-powerful IoT devices while ensuring robustness and generalization. 

System efficiency comes from sensing and data processing. Unlike classical vision systems, neuromorphic systems try to efficiently capture a notion of seeing motion~\cite{roy2019towards, boybat2018neuromorphic, mead2020we, davidson2021comparison, sengupta2019going}. Bio-inspired learning methods, i.e., spiking neural networks (SNNs), address issues related to energy efficiency~\cite{frady2019robust, pang2019fast, rapp2020spiking, liu2014event, schemmel2006implementing, wu2018spatio, hamilton2021best, wozniak2020deep, burr2019role, roy2019towards, schmuker2014neuromorphic, antonik2019human, strukov2019building, zhang2020system}. SNNs have been widely used in many areas of learning and signal processing~\cite{huh2017gradient, neftci2019surrogate, yonekura2020spike, kim2019simple}. These systems have yet to provide robustness and intelligence that matches that from embodied human cognition. For example, the existing bio-inspired method cannot integrate sensory perceptions with actions. SNN applications in machine learning have largely been limited to very shallow neural network architectures for simple problems. Using deep SNN architecture often does not improve learning accuracy and can result in a possible training divergence~\cite{sengupta2019going}. In addition, SNNs lack brain-like robustness and cognitive support.

On the other hand, Hyperdimensional Computing (HDC) is introduced as a promising brain-inspired solution for robust and efficient learning~\cite{kanerva2009hyperdimensional}. HDC is motivated by the understanding that the human brain operates on \textit{high-dimensional} representations of data originated from the large size of brain circuits~\cite{babadi2014sparseness}. It thereby models the human memory using points of a high-dimensional space, that is, with \textit{hypervectors}. HDC performs a learning task after mapping data into high-dimensional space. This encoding is performed using a set of pre-generated \textit{base vectors}. HDC is well suited to address several learning tasks in IoT systems as: (i) HDC is computationally efficient and amenable to hardware level optimization~\cite{imani2021revisiting, karunaratne2020memory, hernandez2021onlinehd}, (ii) it supports single-pass training with no back-propagation or gradient computation, (ii) HDC offers an intuitive and human-interpretable model~\cite{mitrokhin2019learning}, (iii) it is a computational paradigm that can be applied to a wide range of learning and cognitive problems~\cite{moin2020wearable, Poduval2021cognitive, karunaratne2020robust, mitrokhin2019learning, Rasanen2014, rahimi2017high, Jockel2010}, and (iv) it provides strong robustness to noise -- a key strength for IoT systems~\cite{imani2020dual}. Despite the above-listed advantages,  HDC encoding schemes are not designed for handling neuromorphic data. HDC lacks the behavioral resemblance to neurons to extract features from neuromorphic data effectively.

While SNN mimics the physical properties of the brain (how biological neurons are operating), HDC models the brain at a more abstract and functional level. This makes these two computational models complementary. Inspired by the classical and popular memory model, introduced by Atkinson–Shiffrin \cite{10.3389/fphar.2017.00438}, we propose a novel framework that fundamentally combines Spiking neural network and hyperdimensional computing. Our framework, called \Design, enables a scalable and strong cognitive learning system to better mimic brain functionality. \Design creates a cross-layer brain-inspired system that captures information of sensory data from different perspectives: low-level neural activity and pattern-based neural representation. Since both SNN and HDC have memorization capability, they are powerful in preserving spatial and temporal information. Therefore, \Design can ensure advanced learning capability with high accuracy. 
\begin{itemize}[leftmargin=*]
    \item To the best of our knowledge, \Design is the first framework that fundamentally combines SNN and HDC. \Design first exploits a few layers of spiking neural network to extract low-level spatiotemporal information of raw event-based data. Then, it utilizes HDC to operate over SNN output, learn the abstract information, and classifying the data. To ensure robust, efficient, and accurate HDC learning, we present a non-linear neural encoder that transforms data into knowledge at a very low cost and with comparable accuracy to state-of-the-art methods for diverse applications. 
    \item We develop an end-to-end framework that enables co-training of SNN and HDC models. Instead of using deep SNN architecture, we exploit a simple SNN architecture that updates based on gradient rule and connects it to an HDC module capable of fast and single-pass learning. Our framework trains SNN and HDC models simultaneously to ensure that the data generated by SNN is optimal for HDC learning. 
    \item \Design supports online learning from the data stream. In this configuration, we keep the SNN layer static while exploiting HDC single-pass training capability to update the model in real-time. This enables \Design model to learn or update its functionality with very few samples and without paying the cost of storing large-scale train data for iterative learning. 
\end{itemize}

We evaluate \Design on multiple classification problems. Our evaluation shows that \Design provides significant benefits compared to both HDC and SNN architectures: (1) enhance learning capability by exploiting two-stage information processing, and (2) significantly reduces the network size and required parameters to learn complex information. For example, our results indicate that \Design can provide 6.1\% and 3.8\% higher classification accuracy on MNIST and DVS Gesture datasets.

\section{Brain-Inspired Computing Models}
The human brain remains the most sophisticated processing component that has ever existed. The ever-growing research in biological vision, cognitive psychology, and neuroscience has given rise to many concepts that have led to prolific advancement in artificial intelligent accomplishing cognitive tasks~\cite{Lindsay_2020, 10.3389/fnins.2011.00118, 10.3389/frobt.2020.00063}.

\subsection{Analogy from the Brain} \label{subsec:Analogy}
In this work, we enhance the machine learning method by exploring and translating the memory processing capability of the brain~\cite{10.3389/fphar.2017.00438}. To maximize the synergy between anthropogenic concepts and a body \textit{in silico}, we analyzed the distinct neuromorphic nature of SNN and Vector Symbolic Architecture (VSA)~\cite{lee2020neuromorphic, kleyko2021vector}. We found that the two studies approach neuromorphic computing from complementary philosophies: SNN embodies the sensory processing patterns of the brain from a biological standpoint, while the VSA approach processes data from the behavioral patterns. Finally, we evaluate prototypes in both fields using DECOLLE~\cite{10.3389/fnins.2020.00424} (as SNN representative) and Hyperdimensional Computing~\cite{kanerva2009hyperdimensional} (as VSA representative). 

\textbf{Memory model:} The wildly accepted memory model of Atkinson and Shiffrin~\cite{atkinson1968human} includes sensory registers, short-term store, and long-term store. In particular, short-term store (STS) consists of the memory in storage for a short amount of time (referred to as ``short-term memory'') that can be actively engaged in processing (``working memory''). Long-term store (LTS) refers to the memory maintained for long periods of time. Structures centered around the hippocampus serves to process and transfer memory between short-term and long-term storage~\cite{andersen2006hippocampus, olton1979hippocampus}. 

\textbf{SNN (DECOLLE) as STS:}
SNN mimics biological neural networks at the neuronal level, where the representation of the information is the collective state of the spiking neurons, including membrane potential and synaptic states. Given neuromorphic data that are either transformed from frame-based counterparts or captured directly by Dynamic Vision Sensors (DVS), SNN has the structural advantage in accomplishing simple cognitive tasks. In particular, the recurrent nature of DECOLLE renders it ideal for serving both as a sensory processing unit and as an STS. 

\textbf{VSA (HDC) as LTS:}
Hyperdimensional Computing (HDC) mimics the brain on a functional and behavioral level. Just like how the hippocampus represents long-term memory for the sake of efficient storage and retrieval, HDC represents information in hypervectors that can be efficiently and robustly stored in hardware, such as FPGA and GPU, when compared to both non-VSA and most VSA representations. Given informative pieces of data (in this case, the sensory data after being processed by SNN), HDC can efficiently extract higher-level concepts through the process of encoding (what the hippocampus does) and memorization (what the long-term storage does).

\subsection{Hyperdimensional Computing}
The brain's circuits are massive in terms of numbers of neurons and synapses, suggesting that large circuits are fundamental to the brain's computing.
HDC~\cite{kanerva2009hyperdimensional} explores this idea by looking at computing with ultra-wide words -- that is, with very high-dimensional vectors, or hypervectors. The fundamental units of computation in HDC are high dimensional representations of data known as ``hypervectors'', which are constructed from raw signals using an encoding procedure. There exist a huge number of different, nearly orthogonal hypervectors with the dimensionality in the thousands~\cite{kanerva1998encoding}.
This lets us combine such hypervectors into a new hypervector using well-defined vector space operations while keeping the information of the two with high probability.
Hypervectors are holographic and (pseudo)random with i.i.d. components.
A hypervector contains all the information combined and spread across all its components in a full holistic representation so that no element is more responsible for storing any piece of information than another. 

In recent years, HDC has been employed in a range of applications, such as text classification~\cite{kanerva2000random}, activity recognition~\cite{kim2018efficient}, biomedical signal processing~\cite{moin2021wearable}, multimodal sensor fusion~\cite{rasanen2015sequence}, and distributed sensors~\cite{kleyko2014brain, kleyko2018hyperdimensional}.  A key HDC advantage is its training capability in one or few shots, where object categories are learned from one or few examples and in a single pass over the training data instead of many iterations. HDC has achieved comparable to higher accuracy compared to support vector machines (SVMs)~\cite{rahimi2018efficient, imani2019framework}, gradient boosting~\cite{imani2019bric}, and convolutional neural networks (CNNs)~\cite{mitrokhin2019learning}, as well as lower execution energy on embedded processors compared to SVMs~\cite{montagna2018pulp}, CNNs and long short-term memory~\cite{imani2019framework}. 


\subsection{Spiking Neural Network}
Spiking neural networks (SNNs) are brain-inspired solutions for fault-tolerant and energy-efficient signal processing. SNNs take inspiration from the biological functionality of neurons in the human brain to engineering more efficient computing architectures. In the area of machine learning, SNN shares common properties to Recurrent Neural Network (RNNs), such as similarity in general architecture, temporal dynamics, and learning through weight adjustments~\cite{diehl2016conversion}. 
Several works are now establishing formal equivalences between RNNs and networks of spiking leaky integrate-and-fire (LIF) neurons which are widely used in computational neuroscience~\cite{hunsberger2015spiking}. 
In the LIF model, the neuron's state is the momentary activation level that can be pushed higher or lower depending on the incoming spike value. The neuron state will be reset to a lower value after firing the state~\cite{panwar2017arbitrary}.

There are multiple existing update rules in SNN, such as Spike Time Dependent Plasticity (STDP) and gradient descent. STDP depends on pre-synaptic and post-synaptic information~\cite{linares2009memristance}. Several existing hardware solutions have focused on their implementation in the forms of unsupervised and semi-supervised learning~\cite{prezioso2018spike, pedretti2017memristive}. However, these works are limited in learning static patterns or shallow networks. Recent breakthrough research shows that Deep Continuous Local Learning (DECOLLE) provides effective and efficient training with approximate loss that maintains synaptic plasticity~\cite{huh2017gradient, neftci2019surrogate}. 
However, even with a gradient, the ability to train deep SNNs (with several hidden layers) has remained a major obstacle: the inclination for deep SNNs to diverge during training limits its expected potential.

\section{\Design Overview}
In this paper, we propose \Design, the first hybrid solution that fundamentally combines spiking neural networks and hyperdimensional computing. 
Our framework exploits SNN and HDC in the following ways: 
\begin{itemize}
    \item \textbf{Spiking Neural network:} SNN extracts low-level information of neuromorphic data. SNN is like a feature extractor that learns spatiotemporal information of noisy spikes and projects them into meaningful representation. SNN eliminates noisy events that are less frequent in a temporal manner and exploits redundancy to further strengthen spatially correlated information. DECOLLE, in particular, uses deep continuous local learning, where the network errors are computed within each layer, thus requiring little memory overhead for computing gradients.

    \item \textbf{Hyperdimensional Encoding:} HDC performs a higher level learning over spike data generated by SNN. As explained in Section~\ref{sec:HDC}, HDC consists of two layers: encoder and learner. The encoder maps SNN output spikes into high-dimensional space. HDC encoder is unsupervised and significantly efficient since it does not require any training process. Since the encoder is non-linear (Please refer to Section~\ref{sec:HDC} in supplement), a single-layer HDC classifier can effectively learn the data.  The training only operates over the HDC model to keep efficiency and does not propagate to the encoder module. 
\end{itemize}

\begin{figure*}[t!]
    \centerline{
    \epsfig{file=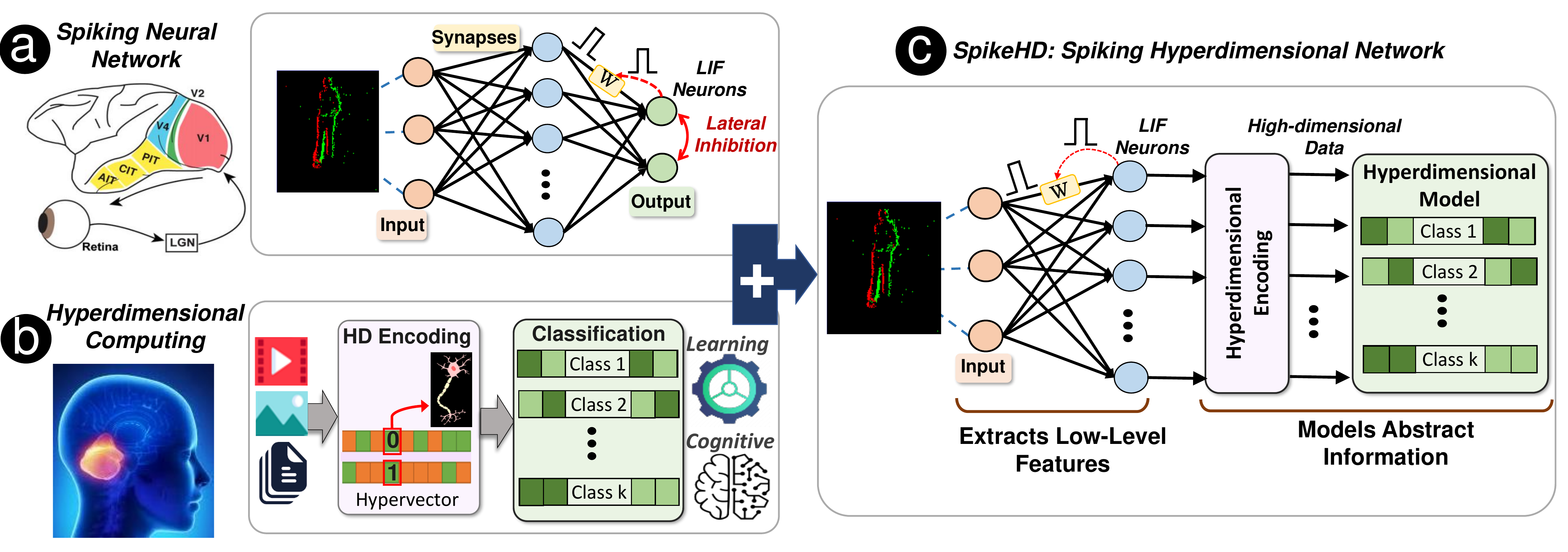, width=1\textwidth}}
     \vspace{-3mm}
    \caption{\Design overview architecture:  spiking hyperdimensional neural network. (a) The leaky-integrate-and-fire (LIF) layers of the SNN have a high synaptic resemblance to the neuronal systems in the brain. This gives rise to its advantage in (the learning of) sensory data processing and maintaining working memory for classification. (b) The high-dimensionality and holography of HDC renders a cerebellum-like functionality with a high capacity memory. (c) the combination of the two as guided by the Atkinson–Shiffrin memory model allows \Design to take advantage of both and overcome their respective shortcomings.}
    \label{fig:spikehd}
\end{figure*}

\textbf{How to Combine SNN \& HDC:} A naive approach can use SNN and HDC in parallel or in series to make a prediction. In the parallel version, both SNN and HDC can make independent predictions, and we can make decisions by looking at the decisions along with their confidence. However, this approach has the following challenges: (1) high computational cost to train two separated models, and (2) decision making and trust in the model is a complex task and requires another learning model.
Similarly, the serial connection of SNN and HDC has the following challenges: (1) an information flow that is limited between the models impedes the ability of both: the latter fails to make good predictions due to a lack of information, and the former fails to be updated due to a lack of loss inferred from the prediction. (2) SNN and HDC are working over different data representations and update rules. SNN works with spike data and is trained using gradient-based rule, while HDC works using dense high-dimensional data representation. This makes HDC and SNN learning not compatible and their interactions non-trivial. 

\textbf{Our Contribution:} To get simultaneous benefits from SNN and HDC, \Design needs to combine them based on their strengths and capabilities. Figure~\ref{fig:spikehd} shows an overview of our hybrid \Design operating over neuromorphic data. \Design exploits two layers of information extraction from event-based sensors: (1) SNN layer to extract low-level information by preserving spatiotemporal correlation of data and (2) hyperdimensional computing to learn the abstract and high-level trend of data. \Design developed a novel framework that co-trains SNN and HDC models. The co-training enables the interaction between SNN and HDC to ensure convergence towards an optimal model. As Figure~\ref{fig:spikehd} shows, hyperdimensional learning, which operates over the spike data, has two components: a non-linear neural encoder that maps SNN output to high-dimensional space and an HDC learning model that combines encoded hypervectors to generate a hypervector representing each class. The HDC model will always take the final prediction. 

In Section~\ref{sec:HDC}, we explained the details of HDC learning. In the rest of the proposal, we present our framework that integrates SNN and HDC (Section~\ref{sec:integration}).

\section{\Design: Integration of Brain Models} \label{sec:integration}
\begin{figure*}[t!]
    \centerline{
    \epsfig{file=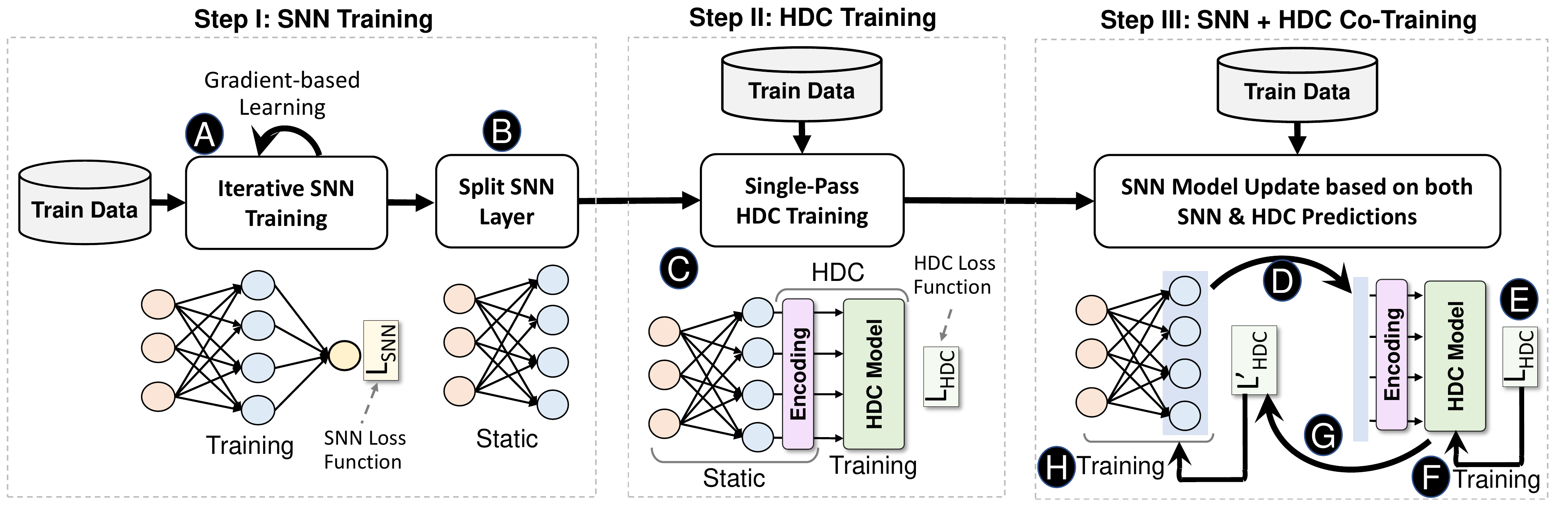, width=1\textwidth}
     }
     \vspace{-3mm}
    \caption{\Designn training process: \textbf{Step I - SNN iterative training.} An original instance of SNN is trained without the influence of HDC, in which initial feature extraction recurrent learning from neuromorphic data are established. \textbf{Step II - HDC single-pass training.} The HDC module is first injected into a deep SNN layer, and train data is propagated from the SNN input layer forward through the HDC module to make a prediction, which leads to modification in the HDC memory component. \textbf{Step III - SNN and HDC co-training.} the loss from the output of HDC module is used to update both HDC memory and SNN layers simultaneously.}
    \label{fig:framework}
\end{figure*}

In this section, we propose a novel framework to combine spiking neural networks and hyperdimensional computing. 
Figure~\ref{fig:framework} shows an overview of our framework combining SNN and HDC. 

\subsection{Step I: SNN Training} 
Our first step aims to establish feature extraction and to fine-tune the short-term memory behavior of the model. The solution starts by training the original SNN model, implemented with DECOLLE, using an entire or a batch of train data. The SNN is a multi-layer network that gets spike data as an input and makes a learning decision on the output layer. Depending on the loss function defined on SNN output, the SNN uses a gradient-based rule to update the synapse's weights ($\invcircledast{A}$). 
During this phase of the training, the SNN learns to extract information from noisy neuromorphic data. Because the synaptic plasticity rules are partially derived from the neural dynamics of the spiking neurons, it is capable of learning spatiotemporal patterns. Increasingly complex features can be learned by adding layers to the SNN, but it is costly and introduces difficulty in convergence. Thus, the number of layers is often limited. 

\subsection{Step II: HDC Training} 
After training the SNN to approximate convergence, the solution applies \textbf{HDC injection}: we split SNN from early layers (often a relatively deep layer) and connect it to the HDC module to extract more abstract information ($\invcircledast{B}$). In this step, the SNN layers are considered to be static because no training happens on SNN. For each train data, we first pass data through SNN. Then, we will be given the SNN representation in the split layer as labeled data to the HDC module for training. 
HDC encoding serves as the ``hippocampus" and maps the spike data into high-dimensional space. Due to the non-linearity of the encoder, an efficient HDC learning module can effectively learn the pattern of data ($\invcircledast{C}$). One advantage of HDC is its capability in learning a one-pass classification model. This eliminates the necessity of a costly iterative method to train a model. As the learning heavily depends on memorization of the encoded hypervectors, the HDC model serves as the long-term memory of the model. 
    
\subsection{Step III: SNN and HDC Co-training} \label{sec:cotraining}
The current approach trains SNN and HDC sequentially, where the SNN training does not get any feedback from HDC module. The SNN module is trained based on the loss function defined at the SNN output layer. However, as we explained, our hybrid architecture performs the prediction using the HDC model. This results in sub-optimal training of \Design architecture. 

To address this issue, we propose a novel co-training method that enables SNN to be trained based on the HDC model prediction. To ensure the SNN layer is well trained for HDC prediction, \Design retrains the SNN layers after HDC training. For every training data or batch of data, \Design starts updating the SNN as follows: 
Firstly, the spike data passes through the split SNN layer ($\invcircledast{D}$) up until the point of injection and generates vectors as input to the HDC module.
Next, the HDC module encodes the input, which is then compared to the HDC model; the loss function is computed against the target label ($\invcircledast{E}$). 
After that, the loss is used to update the HDC model in a single-pass. The learning in HDC is pattern-based and can perform with significantly higher computation efficiency ($\invcircledast{F}$).
Then, HDC back-propagates the loss through the HDC module back to the point of injection ($\invcircledast{G}$). 
Finally, we update the SNN model using a gradient-based rule with the backpropagated loss ($\invcircledast{H}$). 
This procedure continues iteratively over train data until finding a suitable SNN representation that can ensure maximum prediction accuracy in HDC output. 

One thing to note is that the back-propagation in ($\invcircledast{F}$) mainly concerns two matrix multiplications with no significant overhead during training. Passing through the HDC model requires multiplying the loss with the HDC model itself, and passing through the encoder requires the inverse of the activation function and the (Moore-Penrose) inverse of the encoding matrix, the latter of which can be pre-computed upon the initialization of the model. Since the HDC model has faster training than SNN, one can decide to update the HDC model less frequently during the co-training step. During co-training, the HDC loss function can be back-propagated and used to update the SNN layer while the HDC model stays constant. The HDC model update can happen less frequently to ensure lower training costs. 


\subsection{\Designn Online Learning} 
Despite the effectiveness of the proposed framework, the training hybrid SNN and HDC model can be costly for small embedded devices. Our framework requires iteratively repeat co-training phases, which often require a large number of data and iterations to update the quality of the SNN model. Embedded devices may have the following limitations to implement \Design iterative learning: 1) embedded devices often do not have enough memory to keep all train data for iterative learning. As a result, to enable online learning, \Design needs to be updated in one-pass with no need to store train data. 2) embedded devices have limited resources which may not be enough to support the costly gradient-based model update required by SNN.
Here, we propose a solution that enables online \Design learning. As explained in Section~\ref{sec:HDC}, HDC supports single-pass training by creating a model with one-time looking at train data. We exploit this feature to update \Design model in real-time based on the data stream. In this configuration, the SNN model is assumed to be static and does not get updated. Instead, for each batch of data, \Design only updates the HDC model based on the generated loss function. This is similar to transfer learning, where the SNN knowledge is used for new data or environments. 
This model results in much faster convergence and eliminates the necessity of storing train data. 





\subsection{\Design Scalability \& Robustness}
\Design hybrid architecture provides an advanced brain-inspired learning solution with multi-layer information processing. This architecture is significantly strong in preserving spatiotemporal information. \Design also  provides the following advantages: 
\begin{itemize}[leftmargin=*]
    \item \textbf{Scalability:} Using deep SNN architecture often does not improve learning accuracy or results in a possible divergence. \Design hybrid architecture enables SNNs to use effective shallow networks rather than deep non-scalable networks. HDC encoding is used as secondary information processing to provide a high quality of learning while ensuring fast and scalable SNN training.
    \item \textbf{Robustness:} HDC encoding is holographic and redundant, thus provides significant robustness to noise and failure. Our represents stores information of events as a pattern of neural activity in high-dimensional space. Therefore, losing a single or series of dimensions would not remove the information of an event. We further explore on \Design robustness in Section~\ref{sec:robustness}. 
\end{itemize}


\section{Evaluation}
We evaluate the classification performance of \Design on two benchmarks: DVS Gesture Dataset~\cite{amir2017low} and spike-trained MNIST~\cite{lecun-mnisthandwrittendigit-2010}. DVS Gesture Dataset is obtained by Dynamic Vision Sensor (DVS) capturing 11 types of hand and arm gestures from 29 distinct subjects under 3 different lighting conditions. It is thus natively neuromorphic. Spike-trained MNIST, in contrast, is artificial. It comes from processing the original frame-based MNIST images to spike trains, where the serialized pixel values determine the firing rate of the simulated sensors.

The proposed \Design framework has been implemented with two co-designed modules: spiking neural network and hyperdimensional computing. For SNN, we use an existing open-source library~\cite{10.3389/fnins.2020.00424} that trains a network using the gradient-based rule. For HDC, we have developed an in-house library compatible with PyTorch. Our library is an optimized version of PyTorch that better handles the HDC memory requirement for CPU and GPU. 
The default parameters of \Design is as follows. The SNN component consists of 5 LIF layers with respectively $150$, $120$, $100$, $120$, and $150$ neurons. Each LIF layer is associated with a readout layer and a dropout layer as made necessary by local learning. For the HDC component, a dimension of $D=4000$ is used, and the HDC encoder utilizes the hyperbolic tangent function (Tanh) as the activation function. The injection depth that indicates the layer of SNN where HDC is injected is by default 4, which means that it is injected right before the last LIF layer. Finally, the default dataset for evaluation is spike-trained MNIST.

\subsection{Quality of Learning}
Figure~\ref{fig:quality} compares the test classification accuracy of \Design with state-of-the-art SNN. For SNN, we use the conventional DECOLLE architecture in our default configuration. For HDC, we adopt HDC models to directly encode and learn from neuromorphic data with the default dimension and HDC encoder. For different instances of \Design, we apply the default parameters except for our variable - HDC encoder. The models are trained iteratively for 40 epochs, all of which reached convergence. 

Our evaluation shows that default \Design outperforms both SNN and HDC in terms of quality of learning. HDC model alone provides the lowest classification accuracy, as the HDC encoder is weak in extracting spatial and temporal information from noisy spike data. In other words, HDC learning is abstract and cannot be well adapted to extract low-level information from neuromorphic data. In contrast, SNN naturally models the brain's visual systems, thus providing high classification accuracy. However, the SNN accuracy saturates with the increase in the number of layers. In contrast, \Design is a powerful classifier that extracts multi-layer information from the neuromorphic data. Therefore, it eliminates the necessity of using deep SNN layers. Our evaluation shows that \Design achieves, on average, 5.7\% and 3.2\% higher classification accuracy compared to the SNN model after co-training on MNIST and DVS Gesture, respectively.


\subsection{Hyperdimensional Encoding}
\begin{figure*}[t!]
    \centerline{
    \epsfig{file=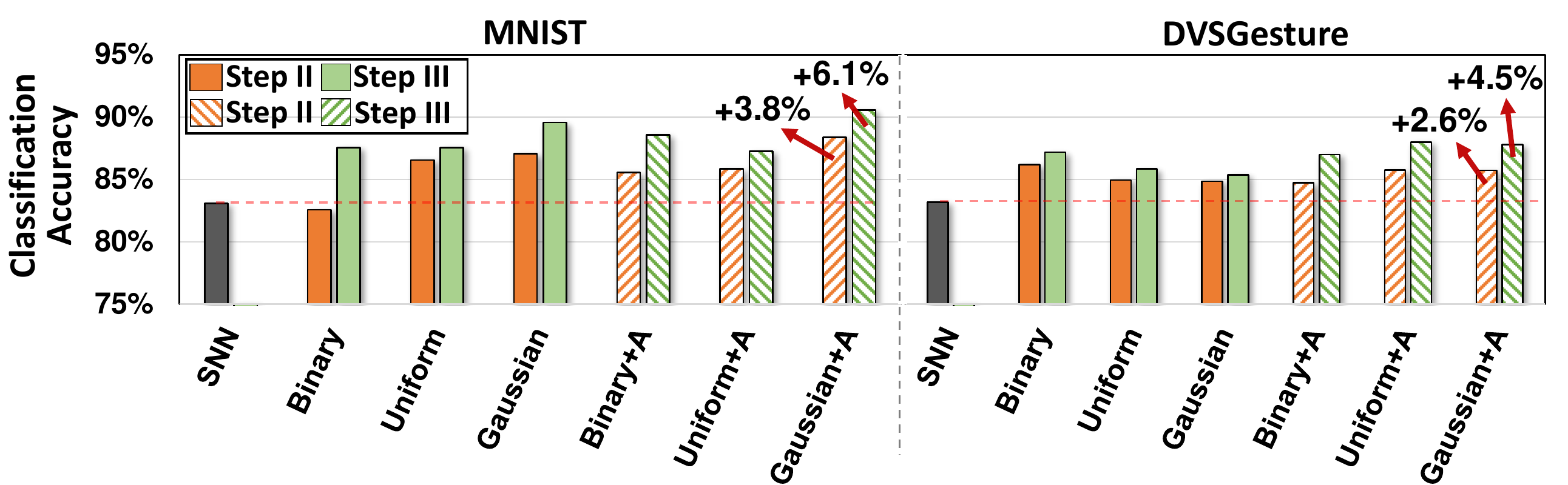, width=0.9\textwidth}
     }
    \caption{Training Performance of \Design with distinct HDC encoders on MNIST and DVSGesture. Each bar plot represent the accuracy of SNN along with \Design using HDC encoders with binary, uniform, or Gaussian base hypervectors. For HDC encoder, the results are reported with and without an activation function (+A indicates an encoder with \textit{Tanh} activation.) }
    \label{fig:quality}
\end{figure*}

To show the impact of our HDC encoder, Figure~\ref{fig:quality} also compares \Design accuracy when HDC is using different encoding modules: linear (Binary)~\cite{SC2021, mitrokhin2019learning}, random projection (Uniform)~\cite{imani2019framework, karunaratne2021robust}, and our proposed non-linear encoder (Gaussian). Our evaluation shows that \Design using a non-linear encoder provides significantly higher classification accuracy compared to the other encoders, independent of the base hypervectors. In particular, linear encoding (labeled as `Binary' in Figure~\ref{fig:quality}) provides the lowest accuracy among them due to limited HDC memory capacity - 4000 bits per class - compared to the ones with non-binary base vectors. The higher accuracy of our encoding comes from: (1) \Design capability in considering the interactions between the features, and (2) exploiting an activation function that makes the mapping non-linear. Our evaluation shows that \Design utilizing non-linear encoder achieves, on average, 3.1\% and 2.4\% higher quality of learning compared to linear and random projection encoder.

Our evaluation also showed that there is progressive improvement in the learning accuracy as the model proceeds with training steps. In most cases, the test accuracy has significant improvement from step~I to step~II, and slightly less from step~III. We will discuss the possible causes in the later evaluations.

\subsection{\Design Training Phases}

\begin{figure*}[t!]
    \centerline{
    \epsfig{file=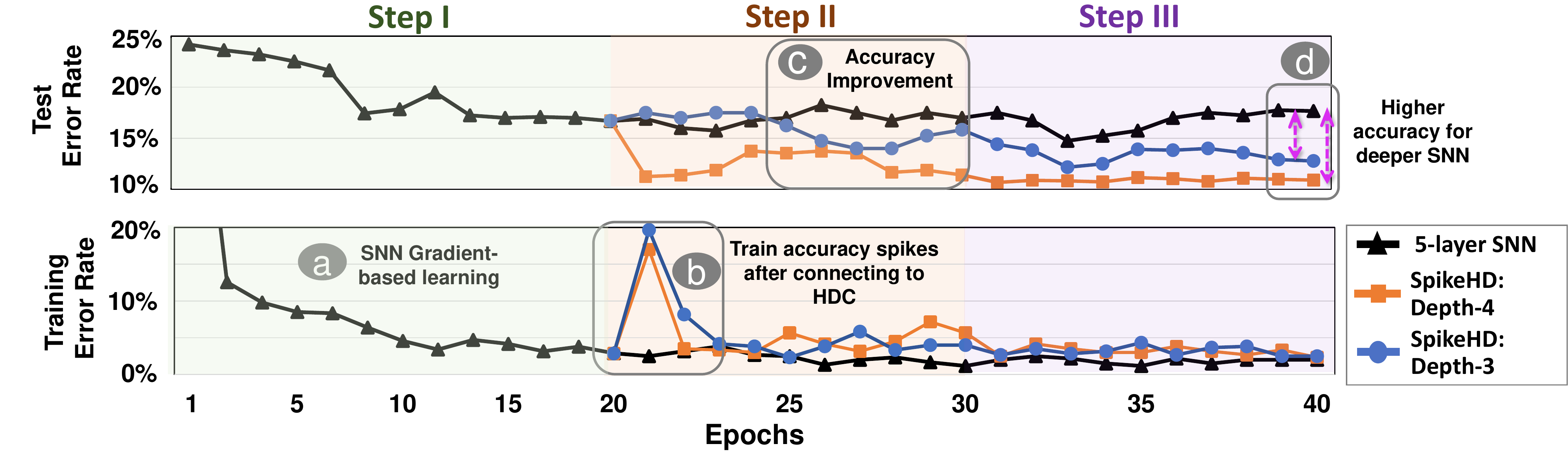, width=0.98\textwidth}
     }
    \caption{Average train and test accuracy of \Design models with a 5-layer DECOLLE and HDC injection at various depths, compared with pure DECOLLE trained for the same number of epochs. Step 1, training SNN only, consists of 20 epochs and reaches model convergence; step 2, training HD module only, and step 3, co-training, both consist of 10 epochs, as convergence is reached earlier in general.}
    \label{fig:spikelayer}
\end{figure*}

Figure~\ref{fig:spikelayer} shows the effect of depth of HDC injection on \Design classification accuracy during both train and test phases. The results are reported for \Design in comparison with DECOLLE with our default setting. 

During step~I of the training (Figure~\ref{fig:spikelayer}a), the accuracy of both the training and testing data quickly increases and stabilizes. This implies the efficiency and effectiveness of DECOLLE in simulating and processing spike data, despite its limited generalization as indicated by the non-trivial difference between test accuracy and train accuracy. 

When the model enters step~II, the training accuracy experience a sharp drop at epoch 20 because we switch from the latter LIF layer to HDC module (Figure~\ref{fig:spikelayer}b). The cause is obvious: since the newly introduced HDC memory is initially a random model, it has no predictive power. That said, the train/test is then reduced and stabilized in about one epoch. In addition, test accuracy experience an improvement (Figure~\ref{fig:spikelayer}c) for the displayed depths of injection, which indicates that HDC modules further extracted and memorized features from the DECOLLE layers that are useful to the prediction. We omitted the result of the trivial model with HDC-injection at depth-5 because it achieves similar performance as the baseline (pure SNN). This is because the depth-5 HDC module connects to the last layer of SNN, which essentially takes the prediction produced by SNN and does learning based on that. 

Finally, the model enters step 3 at epoch 30. Notice that the training accuracy does not have any improvement while the test accuracy has its final improvement (Figure~\ref{fig:spikelayer}d) most accentuated for the depth-3 model. This indicates that the improvement in test prediction comes from the update to the SNN layer in conjunction with that to the HDC memory.

As results indicate, the original SNN network suffers from overfitting, as demonstrated by the difference in train and test accuracy due to the depth of the network. In comparison, \Design mitigates this issue by introducing more explicit memory based on non-linear encoding and is less sensitive to the number of layers. Starting from depth-4, \Design observes significant improvement on the prediction, and the best performance is at depth-4 with the parameters we have chosen; this is partly due to DECOLLE's capacity to extract more meaningful features in the deeper layers, and the long-term memory that the HDC model provides gives rise to increased testing accuracy and, if not eliminated, mitigated the overfitting of DECOLLE model (Figure~\ref{fig:spikelayer}d).

\subsection{\Designn Online Learning}
\begin{figure*}[t!]
    \centerline{
    \epsfig{file=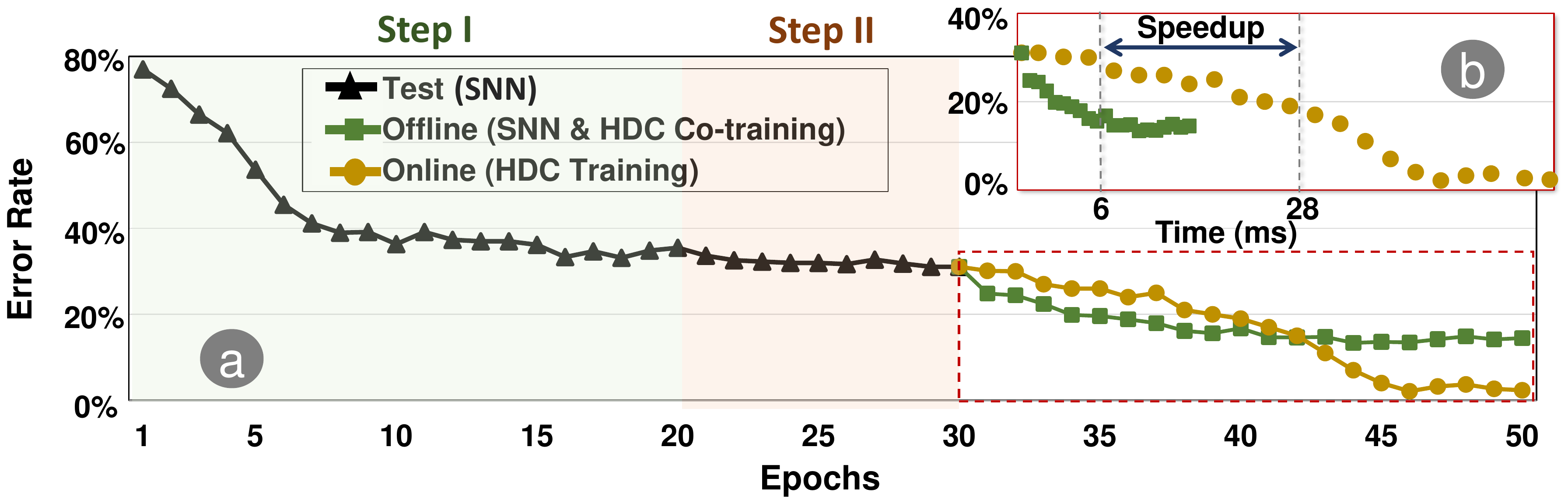, width=0.8\textwidth}
     }
    \vspace{-2mm}
    \caption{Online and offline training performance by epoch and by running time. (a) demonstrates training progress of step~I and II with limited training data (100 samples for MNIST). Then, it is trained with the rest of the data either online, where only the HDC part is updated, or offline, where the co-training happens. (b) demonstrates the training progress measured by running time, with step~II and online/offline step, the former of which is scaled according to training data size for reference.}
    \label{fig:onlineresults}
\end{figure*}

Figure~\ref{fig:onlineresults} compares \Design training efficiency during \Design offline and online training. For fairness, both methods perform Step~I and Step~II \Design training over a small portion of train data. For the rest of the train data, \Design-offline continues updating the model by co-training SNN and HDC (Step~III defined in our framework), while \Design-online only updates the HDC. In other words, \Design-online keeps SNN layers fixed and transfers the learned knowledge for the rest of train data. 


Notice firstly that even with only 100 MNIST samples, 10 for each class, DECOLLE was able to extract many meaningful features. This is implicated by the immediate increase in test accuracy in Step~I (See Figure~\ref{fig:onlineresults}). 
In Step~II, the epoch-wise and time-wise convergence results are reported for both offline and online methods. The time-wise graph is shown as one epoch has different times in offline and online techniques. 
Our results indicate that the online method reaches convergence quicker than the offline method, though the offline method may perform better upon convergence. We observe that the training time of the offline method is much longer than the online learning method. 
This is partly due to DECOLLE's local training. 
We see from the convergence speed and the accuracy improvement that (1) online training incorporates new samples quickly into HDC memory, and (2) the co-training succeeds in backpropagating the loss to the SNN module so that it gets updated for better performance. 

Our evaluation shows that \Design-online can provide comparable accuracy to \Design-offline learning method even when the initial training is very limited (100 samples). In other words, \Design can ignore costly iterative training for a big portion of train data. Instead, it simply updates the HDC model at a minimal cost. 
Our evaluation shows that \Design-online can significantly speedup the training process and also reduces the memory footprint required for training. For example, \Design-online enables 4.6$\times$ faster and 3.1$\times$ lower trainable memory while ensuring the same quality as an online model.

\subsection{\Designn Accuracy and Robustness vs. HDC Dimensionality} \label{sec:robustness}
\begin{figure*}[t!]
    \centerline{
    \epsfig{file=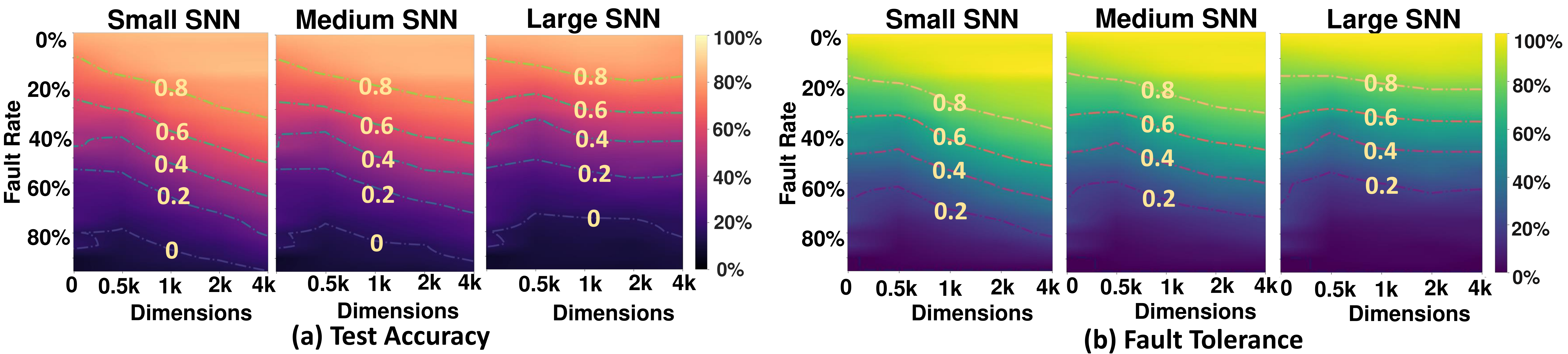, width=1\textwidth}
     }
     \vspace{-1mm}
    \caption{The test accuracy and fault tolerance of \Design under different SNN and HDC parameter settings. (a) Test accuracy is measured by the classification accuracy of the MNIST test data from the model given SNN size, dimension of the HDC memory, and proportion of faulty parameters. (b) Fault Tolerance is measured by the quality of prediction sustained under memory failure, using its 0-fault counterpart and a random classifier (which has a test accuracy of 0.1 in the case of MNIST) as reference.}
    \label{fig:dimension}
\end{figure*}

Dimensionality creates a trade-off between three \Design parameters: accuracy, efficiency, and robustness. Figure~\ref{fig:dimension} shows the impact of hypervector dimension on \Design test accuracy (considering 0\% error rate). \Design in higher dimensionality is a powerful model that can effectively learn the SNN output patterns. \Design provides maximum accuracy even when the dimension reduces from $D=8k$ to $D=4k$. Further decreasing the dimensionality from $D=4k$ results in a minor effect on \Design quality of learning. For example, \Design in $D=2k$ and $D=1k$ provides only 2.3\% and 3.5\% quality loss compared to \Design model in full dimensionality ($D=8k$). 
\Design efficiency also directly depends on the model dimensionality. A higher dimensionality increases the number of required computations in both train and test. However, because the accuracy of HDC modules resembles a sigmoid function along the dimension, to reduce the computation cost, one can decide to use \Design in lower dimensionality. 
For example, reducing \Design dimension from $D=4k$ to $D=2k$ ($D=1k$) results in 1.7$\times$ (3.1$\times$) faster computation. 

We compare \Design computational robustness with SNN. Our evaluation shows that \Design HDC module significantly improves SNN robustness to possible noise and failure. Figure~\ref{fig:dimension} shows \Design accuracy when losing a different proportion of random neurons in the model. The results are reported for \Design using different dimensionality and using different size SNN networks. Our evaluation shows that \Design, in general, provides higher robustness than existing SNN, especially when \Design dimensionality increases. For example, under 10\% random noise, \Design and SNN maintain 94.0\% and 87.1\% quality. 
The ability to sustain prediction quality generally increases as the dimension of HDC memory increases, though it does generate a slight dip when the HDC dimension is low. This can be attributed to the fault tolerance of DECOLLE and the higher vulnerability of low-dimensional HDC modules. 
Our results indicate that test accuracy has only a slight advantage when the HDC dimension is high, and SNN is large. This advantage will be more accentuated in smaller SNN, or in more complex tasks.

\vspace{-1mm}
\section{Conclusion and Discussion}
\begin{figure*}[t!]
    \centerline{
    \epsfig{file=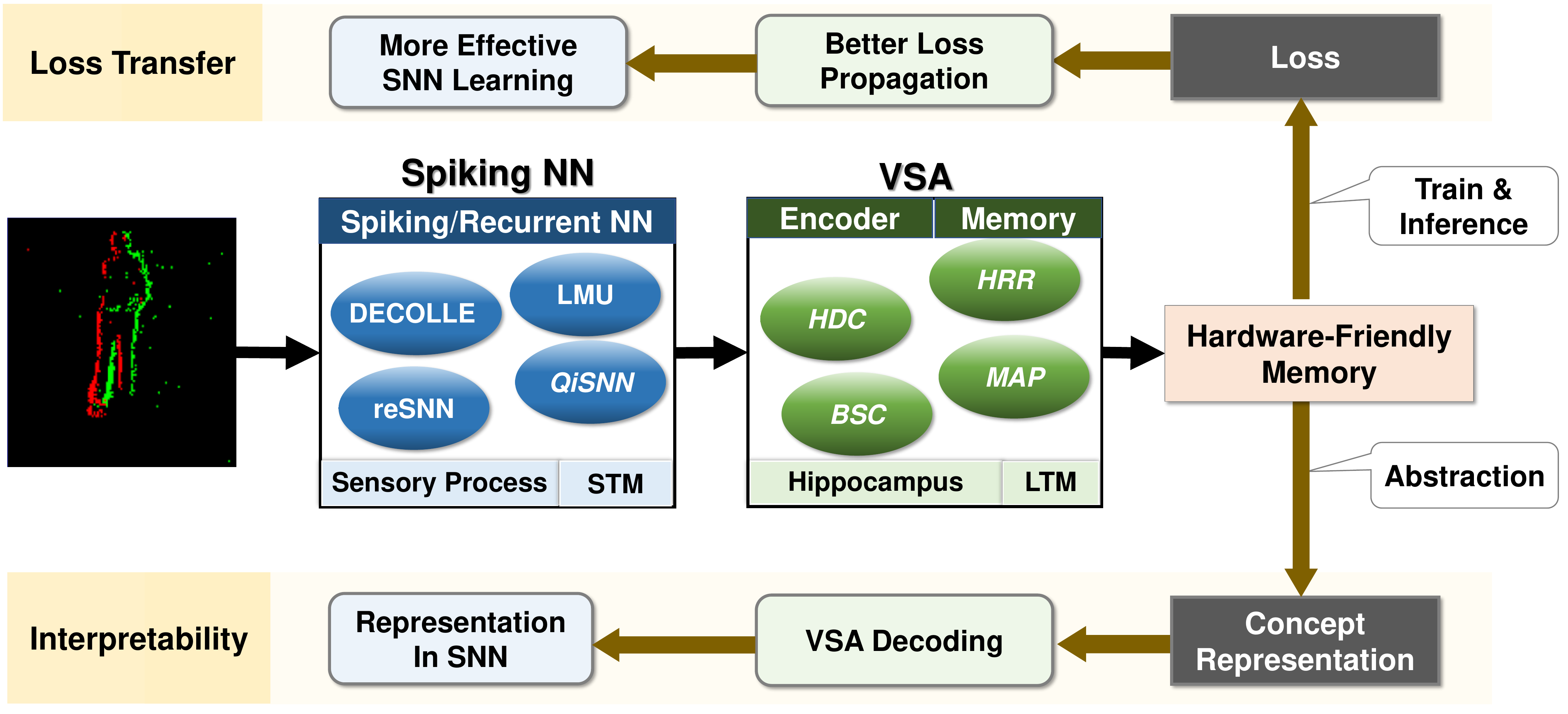, width=0.9\textwidth}
     }
     \vspace{-3mm}
    \caption{Overview of \Design extended. \textbf{Framework}: \Design incorporates spiking neural networks and vector symbolic architecture at a high level. While our work has demonstrated one efficiency-oriented instance of the model with DECOLLE and HDC, other combinations may give rise to models of different strengths. \textbf{Loss Transfer}: one direction for future work is the back-propagation of loss from VSA. Better loss propagation to the SNN layer leads to more effective SNN training. \textbf{Interpretability}: The decoding of HDC Memory and the interpretability of SNN lead to knowledge at the sensory data level.}
    \label{fig:discuss}
\end{figure*}

In this paper, we propose \Design, a novel framework that combines Spiking neural network and hyperdimensional computing in order to design a scalable and strong cognitive learning system that better mimics brain functionality. 
\Design exploits spiking neural networks to extract low-level features by preserving the spatial and temporal correlation of raw event-based spike data. Then, we utilize HDC to operate over SNN output by mapping the signal into high-dimensional space, learning the abstract information, and classifying the data. 
Our evaluation on a wide range of classification problems shows that \Design provides significant benefit compared to both HDC and SNN architecture: (1) enhances learning capability by exploiting two-stage information processing, (2) significantly reduces the network size and required parameters to learn complex information. For the rest of this section, we highlight some of the open challenges that our framework has yet to overcome and encourage exploration of the question along multiple axes (Figure~\ref{fig:discuss}).

\textbf{Loss Backpropagation:}
During step~III of \Design, a Moore-Penrose inverse of the HDC encoder is applied to backpropagate the loss from the HDC module to the SNN. Since HDC encoder maps vectors to hypervectors, the rank of the inverse is limited to the output dimension of the SNN at the point of injection, which may be way less than the dimension. A large amount of information may be lost from this transition. We experimented with several methods to solve this problem. One example is to continue training the original SNN, transfer the new weights to the \Design, and then train the HDC module. This did not improve performance except in the case of the transfer learning task, where the context of the data or the task changes. One method that may be suggested is to introduce a regularization term in the loss function of the SNN layers such that it outputs an HDC-like vector as the representation of the data directly. This will avoid the explicit usage of the HDC non-linear encoder, and the loss will be optimally propagated up to the approximation introduced by the regularization term.

\textbf{Component Choices:}
We have selected DECOLLE as SNN and HDC as VSA for our hybrid model for the reason we've discussed in section\ref{subsec:Analogy}. It is optimized for time and energy efficiency, and practicality, as both models are known for such traits. Readers interested in the exploration of other aspects may choose to adopt our memory framework to other components. such as Legendre Memory Units~\cite{Voelker2019LegendreMU} and HDC, or BI-SNN and HRR~\cite{plate1995holographic}. 

\textbf{Concept Interpretability:}
Our current usage of the memory framework is to directly operate on long term memory and derives decisions from its representation, which simulates what the cerebellum does. For the purpose of completing the analogy to the Atkinson-Shiffrin memory model, it has yet to be incorporated the decoding mechanisms of the HDC memory: we did not fetch the long term memory back to the hippocampus and decode it for operations in working memory. This subject is not the purpose of this paper, and the decoupling of encoding and decoding invites more possibilities, as the HDC memory may be used as heterogeneous storage such that multiple tasks may be performed in one memory model.

That said, the subject of interpretability remains interesting at two levels. Each entry in the HDC memory represents the concept of the class, which can be decoded to retrieve a representation of the concept at the SNN level. The representation in SNN can then be interpreted to infer knowledge on the original sensory data.

\section*{Acknowledgment}
This work was supported in part by National Science Foundation (NSF) \#2127780, Semiconductor Research Corporation (SRC) Task No. 2988.001, Department of the Navy, Office of Naval Research, grant \#N00014-21-1-2225, and generous gifts from Cisco.

\ifx{
\subsection{Training modularity}
\Design takes a step-by-step training approach that comes with both advantages and limitation. For the advantages, independently training each modules allows guided training that facilitates task assignment of the components. We trained the SNN module so that its feature extraction ability for spatio-temporal representations goes unimpeded, maximizing its intended usage as a processor that transform neuromorphic data to a representation that is more \textit{silico}-friendly. The HDC model is trained right after to establish the long term memory of the model, whose representation is encoded by HD encoder to hyperdimensional space, which leads to more efficient memorization and concept learning. Finally, the co-training of SNN and HD improves the synergy between the two, as they have established their roles and will not deviate from them. The disadvantage of the model, to speak from a neural network standpoint, is that the training and the model is not homogeneous. There are multiple steps involved in training, and this introduces more hyperparameters for fine-tuning a model (when should we move to the next step?). The training of HDC and SNN are essentially separate (though in parallel), and their cooperation introduces even more hyperparameters (what's a good learning rate ratios for HDC and SNN in step 3?). 

To circumvent such issue, one idea would be to discover mechanisms that ``auto-differentiate'' the roles of SNN and HDC. Neural Turing Machine~\cite{graves2014neural} is an excellent example of this. It has an explicit memory component and is accessed via  read/write parameters generated from the controller, an recurrent or convolutional neural network. There are many challenges in operating on HDC memory due to its hyperdimension and unique operations, mainly binding and bundling. To the best of our knowledge, the only published work relating to this lead is ~\cite{Karunaratne_2021}, which nonetheless use HDC memory naively without any bundling or binding. This topic remains an active research area in BIASLab at UCI. 
}\fi 

\vspace{-1mm}
{\small 
\bibliographystyle{ieeetr}
\bibliography{mybibliography}
}

\newpage
\setcounter{equation}{0}
\setcounter{section}{0}
\setcounter{figure}{0}
\setcounter{table}{0}
\setcounter{page}{1}
\renewcommand{\theequation}{S\arabic{equation}}
\renewcommand{\thefigure}{S\arabic{figure}}

\setcounter{page}{1} 

 {\bf \Large
Supplementary material for ``Spiking Hyperdimensional Network: Neuromorphic Models Integrated with Memory-Inspired Framework"}

\vspace{3mm}

\section{Hyperdimensional Classification} \label{sec:HDC}
We present a robust and lightweight hyperdimensional classification. Figure~\ref{fig:classification}a shows an overview of HDC learning. 
The first step in HDC is to encode data into high-dimensional space. Then, HDC performs a learning task over encoder data by performing a single-pass training. The result of training will be to generate a hypervector representing each class. The inference task can perform by checking the similarity of an encoded query to the class hypervector.

\subsection{Non-Linear Encoding}
\label{sec:HDencoding}
The encoder is the most important component of hyperdimensional learning. The goal of the encoder is to map data into high-dimensional space such that we can extract knowledge from data at a lower cost. The encoder selection depends on the data type (structure of data) and the target learning task. 
There are multiple existing HDC-based encoding methods~\cite{rahimi2017hyperdimensional2, mitrokhin2019learning, nazemi2020synergiclearning}.
Although these methods have shown excellent classification accuracy for their application-specific problems,
to the best of our knowledge, the existing encoding methods linearly combine the hypervectors corresponding to each feature, resulting in sub-optimal classification quality for general classification problems.
To obtain the most informative hypervectors, the HDC encoding should consider the non-linear interactions between the feature values with different weights.

In this context, we propose a novel encoding method that exploits the kernel trick~\cite{rahimi2008random, scholkopf2001kernel} to map data points into the high-dimensional space.
The underlying idea of the kernel trick is that data, which is not linearly separable in original dimensions, might be linearly separable in higher dimensions.
Let us consider certain functions $K(x,y)$ which are equivalent to the dot product in a different space, such that $K(x,y) = \Phi(x) \cdot \Phi(y)$, where $\Phi(\cdot)$ is often a function for high dimensional projection. 
The Radial Basis Function or Gaussian Kernel is the most popular kernel:
\vspace{-1.5mm}
$$K(x,y) = e^{\frac{-||x - y||^2}{2\sigma^2}}$$

\noindent We can take advantage of this implicit mapping by replacing their decision function with a weighted sum of kernels: 
\vspace{-2.5mm}
$$f(\cdot) = \sum_{i = 0}^{N} c_i K(\cdot,x_i) $$ 
 
\noindent where $(x_i,y_i)$ is the training data sample, and the $c_i$s are constant weights.
The study in~\cite{rahimi2008random} showed that the inner product can efficiently approximate Radial Basis Function (RBF) kernel, such that:
$$K(x,y) = \Phi(x) \cdot \Phi(y) \approx z(x) \cdot z(y) $$

\noindent The Gaussian kernel function can now be approximated by the dot product of two vectors, $z(x)$ and $z(y)$.

\begin{figure*}[t!]
    \centering
    \epsfig{file=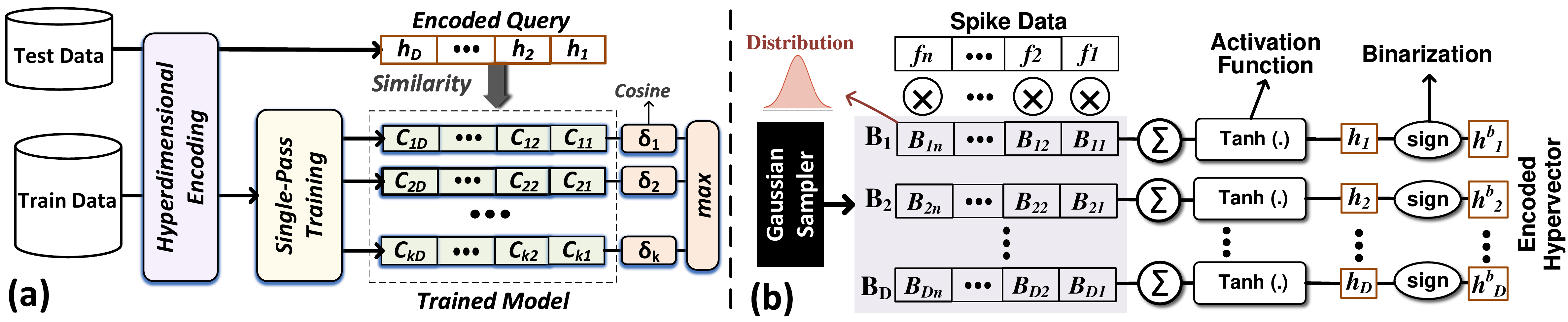, width=1\textwidth}
    \vspace{-6mm}
    \caption{(a) Overview of hyperdimensional classification during both training and testing phases, and (2) HDC non-linear encoder.}
    \label{fig:classification}
    \vspace{-2mm}
\end{figure*}

The proposed encoding method is inspired by the RBF kernel trick method. Figure~\ref{fig:classification}b shows our encoding procedure.
Assume an input vector in original space $\vec{{F}} = \{f_1,~f_2,\cdots,f_n\}$ and $F \in \mathcal{R}^n$. The encoding module maps this vector into high-dimensional vector, $H = \{h_1,~h_2,\cdots,h_D\} \in \mathcal{R}^D$, where $D \gg n$. The following equation shows an encoding method that maps input vector into high-dimensional space:
\begin{equation}
h_i = \cos(\vec{F} \cdot \vec{\mathcal{B}}_i + b_i) \sin(\vec{F} \cdot \vec{\mathcal{B}}_i)
\end{equation}

where $\vec{\mathcal{B}}_{k}$s are randomly chosen hence orthogonal base hypervectors of dimension ${D} \simeq 10k$ to retain the spatial or temporal location of features in an input and $b_i \sim \mathcal{U}(0, 2\pi)$. That is, $\vec{\mathcal{B}}_{kj} \sim \mathcal{N}(0, 1)$ and $\delta(\vec{\mathcal{B}}_{k_1}, \vec{\mathcal{B}}_{k_2}) \simeq 0$, where $\delta$ denotes the cosine similarity. However, this activation is not a convex function, thus making it impossible to back-propagate from HDC encoder. Although HDC learning does not rely on back-propagation, in section~\ref{sec:cotraining}, we will show that the integration of SNN and HDC requires back-propagation through HDC encoding module. To address this issue, we exploit hyperbolic tangent function (Tanh) as an activation function: 

\begin{equation}
h_i = \tanh{(\vec{F} \cdot \vec{\mathcal{B}}_i + b_i)}
\end{equation}

After this step, each element $h_i$ of a hypervector $\mathbf{H}^n$ has a non-binary value.
In the HDC, binary (bipolar) hypervectors are often used for computation efficiency.
We thus obtain the final encoded hypervector by binarizing it with a sign function ($\mathbf{H}=sign (\mathbf{H}^n)$) where the sign function assigns all positive hypervector dimensions to `1' and zero/negative dimensions to `-1'. The encoded hypervector stores the information of each original data point with $D$ bits.
In our example, the feature vector, $F$, is highly sparse event-based data. However, after mapping data through the above encoder, the generated high-dimensional data will get a dense binary representation. 


\subsection{Hyperdimensional Model Training}
We exploit hyperdimensional learning to directly operate over encoded data. Figure~\ref{fig:classification}a shows an overview of HDC classification. 
HDC identifies common patterns during learning and eliminates the saturation of the class hypervectors during single-pass training. Instead of naively combining all encoded data, our approach adds each encoded data to class hypervectors depending on how much new information the pattern adds to class hypervectors. If a data point already exists in a class hypervector, HDC will add no or a tiny portion of data to the model to prevent hypervector saturation. If the prediction matches the expected output, no update will be made to avoid overfitting. This adaptive update provides a higher chance and weight to non-common patterns to represent the final model. This method can eliminate the necessity of using costly iterative training.


Let us assume $\vec{\mathcal{H}}$ as a new training data point. HDC computes the cosine similarity of $\vec{\mathcal{H}}$ with all class hypervectors. We compute similarity of a data point with $\vec{\mathcal{C}}_i$ as: $\delta(\vec{\mathcal{H}},  \vec{\mathcal{C}}_i)$. Instead of naively adding a data point to the model, HDC updates the model based on the $\delta$ similarity. 
If an input data has label $l$ and correctly matches with the class, the model updates as follows:
\begin{equation}\label{eq:update}
\begin{split}
\vec{\mathcal{C}_l} \gets  \vec{\mathcal{C}_l} + \eta_1~(1-\delta_{l}) \times \mathcal{\vec{H}}  \\
\end{split}
\vspace{-4mm}
\end{equation}

\noindent where $\eta$ is a learning rate. A large $\delta_{l}$ indicates that the input is a common data point which is already exist in the model. Therefore, our update adds a very small portion of encoded query to model to eliminate model saturation ($1-\delta_{l} \simeq 0$). If the input data get an incorrect label of $l'$, the model updates as:

$$\vec{\mathcal{C}_{l'}} \gets  \vec{\mathcal{C}_{l'}} - \eta_2~(\delta_{l'}-\delta_{l}) \times \mathcal{\vec{H}} $$

\noindent where $\delta_{l'}-\delta_{l}$ determines the weight that the model needs to be updated. Small $\delta_{l'}-\delta_{l}$ indicates that the query is marginally mismatched while larger mismatch is updated with a larger factor ($\delta_{l'}-\delta_{l} \gg 0$). 

\ifx{
\subsection{HDC Iterative Learning}
HDC has an option of supporting retraining to enhance the quality of the model. HDC retraining starts from the initial adaptive model (explained in Section~\ref{sec:singlepass}). \Design initial model already considered the weight of each input data during single-pass training. 
This enables HDC to retrain the model with a very small number of iterations, resulting in fast convergence. 
Figure~\ref{fig:overview} shows \Design functionality during adaptive retraining. 
\Design follows a similar learning procedure as initial training. For each training data point, say $\vec{\mathcal{H}}$, \Design checks the similarity of data with all class hypervectors in the model ($\invcircledast{A}$) and updates the model for each miss-prediction ($\invcircledast{B}$). 
Retraining examines if the model correctly returns the label $l$ for an encoded query $\vec{\mathcal{H}}$. If the model mispredicts it as label $l'$, the model updates as follows ($\invcircledast{C}$).
\begin{equation}\label{eq:update}
\begin{split}
\vec{\mathcal{C}_l} \gets \vec{\mathcal{C}_l} + \eta~ (1-\delta_{l}) \times \mathcal{\vec{H}} \\
\vec{\mathcal{C}_{l'}} \gets \vec{\mathcal{C}_{l'}} - \eta~ (1 - \delta_{l'}) \times \mathcal{\vec{H}}
\end{split}
\vspace{-3mm}
\end{equation}

\noindent where $\delta_l = \delta(H,\vec{\mathcal{C}_{l}})$ and $\delta_{l'} = \delta(H,\vec{\mathcal{C}_{l'}})$ are the similarity of data with correct and miss-predicted classes, respectively. This ensures that we update the model based on how far a train data point is miss-classified with the current model. In case of of a very far miss-prediction, $\delta_{l} \ll 0$, \Design retraining makes a major changes on the model. While in case of marginal miss-prediction, $\delta_{l} \simeq 0$, the update makes smaller changes on the model.
We also provide separate coefficients for the true and miss-predicted labels, allowing \Design to update each class hypervector independently.
}\fi

\subsection{Hyperdimensional Inference}
In inference, HDC checks the similarity of each encoded test data with the class hypervector in two steps. The first step encodes the input (the same encoding used for training) to produce a query hypervector $\vec{\mathcal{H}}$. Then we compute the similarity ($\delta$) of $\vec{\mathcal{H}}$ and all class hypervectors. Query data gets the label of the class with the highest similarity.

\section{\Designn Trainable parameters} \label{sec:paramsize}

\begin{figure}[t!] 
    \centerline{
    \epsfig{file=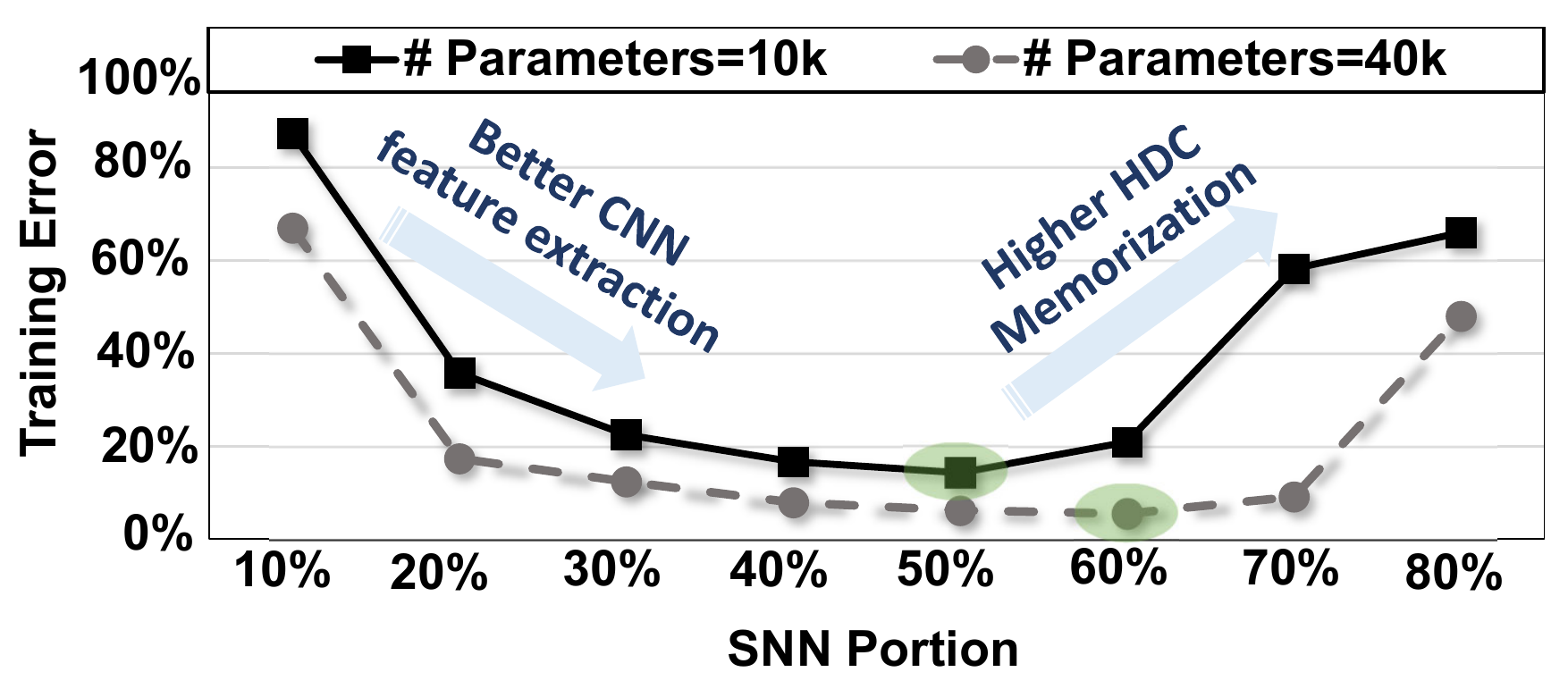, width=0.6\columnwidth}
     }
    \caption{\Design performance with different SNN/HDC training parameter ratios. The lines describe the average training accuracy of \Design varying on the amount of trainable parameters that is assigned to SNN.}
    \label{fig:SNNportion}
\end{figure}

Figure~\ref{fig:SNNportion} demonstrates the effect of trainable parameter ratio on the performance of the model. To stress-test our model, the shape of the SNN is limited to 3 LIF layers with size ratio 2:3:2, and the dimension of HDC is derived from however many trainable parameter is allowed in HDC Memory, up to residue. For small \Design (blue line with 10k parameters), there is a fine spot where optimal accuracy is reached. This is due to the fact that the training data requires an SNN big enough to extract meaningfull features (which explain the left slope) and an HDC memory large enough to memorize these features (which explain the right slope). Due to the fact that HDC memory accuracy resembles a sigmoid function with respect to dimension in practice, the accuracy deteriorate rapidly upon lowering the dimension to a certain threshold. This is demonstrated in the rapid increase in error rate for \Design with very high SNN proportion. In addition, due to the correlation of the dimension threshold to the output dimension of the SNN instead of the size of the SNN, large \Design (black line with 40k parameters) requires smaller proportion of HDC memory to perform well. The demand for the size of HDC also comes the number of classes. For example, the Omniglot dataset, a few-shot, 1623-class classification task, demands high HDC memory.

\end{document}